\documentclass[runningheads]{llncs}
\usepackage[T1]{fontenc}
\usepackage{graphicx}
\usepackage{float} 
\usepackage{tabularx}
\usepackage[table]{xcolor}
\begin{document}
\title{CODE-GEN: A Human-in-the-Loop RAG-Based Agentic AI System for Multiple-Choice Question Generation
}
\titlerunning{CODE-GEN: An agentic AI system for MCQ Generation}
\author{Xiaojing Duan\inst{1}\orcidID{0000-0003-0822-2022} \and
Frederick Nwanganga\inst{1}\orcidID{0000-0002-4514-6527} \and
Chaoli Wang\inst{1}\orcidID{0000-0002-0859-3619}}
\authorrunning{X. Duan et al.}
\institute{University of Notre Dame, Notre Dame IN 46556, USA\\ 
\email{\{xduan,fnwangan,chaoli.wang\}@nd.edu}
}

\maketitle              % typeset the header of the contribution
%/
\begin{abstract}
We present CODE-GEN, a human-in-the-Loop, retrieval-augmented generation (RAG)-based agentic AI system for generating context-aligned multiple-choice questions to develop learners’ code reasoning and comprehension abilities. CODE-GEN (Context-aligned, Output-validated, Dual-agent, Expert-guided GENeration) employs an agentic AI architecture in which a Generator agent produces multiple-choice coding comprehension questions aligned with course-specific learning objectives, while a Validator agent independently assesses content quality across seven pedagogical dimensions. Both agents are augmented with specialized tools that enhance computational accuracy and verify code outputs. To evaluate CODE-GEN’s performance, we conducted an evaluation study involving six human subject-matter experts (SMEs) who judged 288 AI-generated questions. The SMEs produced a total of 2,016 human-AI rating pairs, indicating agreement or disagreement with the Validator’s assessments, along with 131 instances of qualitative feedback. Analyses of SME judgments show strong system performance, with human-validated success rates ranging from 79.9\% to 98.6\% across the seven pedagogical dimensions. The analysis of qualitative feedback reveals that CODE-GEN achieves high reliability on dimensions well suited to computational verification and explicit criteria matching, including question clarity, code validity, concept alignment, and correct answer validity. In contrast, human expertise remains essential for dimensions requiring deeper instructional judgment, such as designing pedagogically meaningful distractors and providing high-quality feedback that reinforces understanding. These findings inform the strategic allocation of human and AI effort in AI-assisted educational content generation.

% \keywords{Human-in-the-loop evaluation \and AI-assisted assessment \and Agentic AI \and Programming education \and Assessment quality \and Pedagogical alignment \and Large language models \and Retrieval-augmented generation}
\end{abstract}
\newpage
\section{Introduction}
Recent advances in Large Language Models (LLMs) have generated substantial interest in their potential to support teaching and learning in educational contexts 
%ogunleyeSystematicReviewGenerative2024,
\cite{dutulescuYMCQReasoningEnhancedMCQ2025,scariaAutomatedEducationalQuestion2024}. While prior studies demonstrate that LLMs can generate a wide range of educational artifacts
%juryEvaluatingLLMgeneratedWorked2024,
\cite{hoqFacilitatingInstructorsLLMCollaboration2025,barrosLargeLanguageModels2025}, their adoption in authentic classroom settings remains limited \cite{kurdiSystematicReviewAutomatic2020,lyuEvaluatingEffectivenessLLMs2024}. This limitation stems from two persistent challenges. First, content produced by general-purpose LLMs is often overly generic and insufficiently aligned with course-specific learning objectives \cite{kasneciChatGPTGoodOpportunities2023,pintoLessonsBuildingStackSpot2024}, undermining instructional validity and reducing instructor trust in AI-assisted content generation. Second, although recent work in agentic AI has introduced multi-agent systems with specialized roles for generation, critique, and revision to improve content quality \cite{yukselMultiAIAgentSystem2024,kostopoulosAgenticAIEducation2025}, these systems frequently assume that automated critique agents provide reliable evaluations. In practice, such assumptions are problematic, as LLM-based evaluators are known to exhibit hallucinations, bias, and inconsistent judgment \cite{stureborgLargeLanguageModels2024,linLLMbasedAgentsSuffer2025}. Without systematic validation against human expert judgment, errors introduced at the evaluation stage risk being amplified rather than corrected.

To address these gaps, this study presents CODE-GEN (Context-aligned, Output-validated, Dual-agent, Expert-guided GENeration), an agentic AI system designed to 
generate contextually grounded multiple-choice coding comprehension questions. CODE-GEN integrates retrieval-augmented generation (RAG) with a dual-agent architecture in which a Generator agent produces multiple-choice coding questions aligned with course-specific learning objectives, and a Validator agent independently assesses question quality across seven pedagogical dimensions derived from established multiple-choice item-writing guidelines. Both agents are augmented with specialized tools to support computational accuracy and reliable code execution verification. To evaluate the effectiveness of this agentic approach, we conducted a comprehensive human evaluation study in which six subject-matter experts (SMEs) judged 288 AI-generated questions, generated 2,016 human-AI judgment pairs, and provided 131 instances of qualitative feedback. 
% Each question was accompanied by the Validator agent’s dimension-level assessments based on established multiple-choice item-writing guidelines. 
%In total, the experts provided 2,016 human-AI judgment pairs and 131 instances of qualitative feedback.
Analysis of these SME judgments provides empirical evidence of where agentic AI systems can reliably support educational content generation and where human expertise remains essential, thereby characterizing both the strengths and limitations of LLM-based approaches to AI-assisted educational content generation.

The main contributions of the study are as follows: (1) We introduce CODE-GEN \footnote{All resources associated with CODE-GEN will be made available upon request.} as a Human-in-the-Loop, agentic AI system for generating high-quality multiple-choice coding comprehension questions that are context-aligned, pedagogically grounded, and systematically validated against explicit quality criteria; (2) We conduct a rigorous human expert evaluation of CODE-GEN, using SME judgments to evaluate the reliability of automated validation across seven pedagogical dimensions; (3) We derive evidence-based insights into the division of labor between AI and human expertise in educational content generation, clarifying which aspects can be reliably automated and which require deeper instructional judgment, particularly for pedagogical depth and targeting student misconceptions.

\section{Related Work}
% Prior research has demonstrated the potential of large language models to support automated question and assessment generation across a range of educational domains. However, two strands of work are particularly relevant to the present study: retrieval-augmented generation for grounding model outputs in instructional context, and agentic AI approaches that distribute generation and evaluation across multiple coordinated agents.

\subsection{RAG}
RAG has emerged as an effective technique for constraining LLM outputs using domain-specific knowledge sources \cite{hanImprovingAssessmentTutoring2024}. %,parvezRetrievalAugmentedCode2021,zhaoRetrievalAugmentedGenerationAIGenerated2024}. 
In educational settings, studies show that RAG-based systems produce assessment artifacts with greater contextual specificity, improved alignment with learning objectives, and fewer factual errors than standalone LLM generation \cite{mittalAskademiaRealTimeAI2025}.
%,maityLeveragingInContextLearning2024,levonianRetrievalaugmentedGenerationImprove2023}. 
In a typical RAG pipeline, relevant instructional materials are retrieved from external repositories and incorporated into the model’s prompt to guide generation \cite{lewisRetrievalAugmentedGenerationKnowledgeIntensive2020}. By conditioning outputs on retrieved examples, RAG limits the model’s tendency to generate generic or decontextualized questions. It helps ensure that generated assessments reflect the intended instructional context %\cite{dongHowBuildAdaptive2023,levonianRetrievalaugmentedGenerationImprove2023,
\cite{nanongkhaiEvaluatingAdaptiveGenerative2026}. Collectively, this work suggests that RAG is a promising foundation for AI-assisted assessment generation in authentic classroom settings. The present study builds on this literature by embedding RAG directly into an agentic architecture designed not only to generate, but also to evaluate assessment quality.

\subsection{Agentic AI}
Recent advances in agentic AI have motivated the use of multi-agent systems for educational applications \cite{caoFirstDraftFinal2025,houLLMEnhancedMultiagentArchitecture2025}. These systems typically assign distinct roles to different agents, enabling workflows in which one agent generates content and another critiques or revises it \cite{russellArtificialIntelligenceModern2016,acharyaAgenticAIAutonomous2025}. Empirical studies report improvements in fluency, diversity, and coverage of generated assessment items relative to single-pass LLM outputs \cite{kostopoulosAgenticAIEducation2025,kamalovEvolutionAIEducation2025}.
% ,artsinCHARTINGNEWHORIZONS2025}. 
A common design pattern is the use of critique or evaluator agents whose feedback refines the generated questions \cite{qianDeanLLMTutors2025,wangGeneratingAILiteracy2024}. However, this line of work often assumes that automated evaluators provide reliable and pedagogically sound judgments. Recent evidence challenges this assumption, showing that LLM-based evaluators can exhibit hallucinations, bias, and inconsistent reasoning 
%\cite{stureborgLargeLanguageModels2024,
\cite{linLLMbasedAgentsSuffer2025,zhangMIRAGEBenchLLMAgent2025}. When evaluator outputs are treated as ground truth and reused recursively, such errors risk being amplified rather than corrected. Despite these risks, few studies systematically validate automated evaluators' performance against human expert judgment, limiting the trustworthiness of agentic assessment-generation systems in high-stakes educational contexts. To bridge this gap, the present study explicitly evaluates the critique agent's performance against human SMEs, then uses its feedback to drive regeneration or improvement. By situating the critique agent's performance as an empirical object of study rather than an assumed capability, this work contributes a human-centered perspective on how agentic AI can be used to support high-quality educational content generation while clarifying the boundaries of reliable automation.

\section{Method}
\subsection{CODE-GEN}
CODE-GEN is a human-in-the-loop, agentic AI system designed to generate and validate multiple-choice coding comprehension questions that are closely aligned with course-specific learning objectives. The system integrates RAG with a dual-agent architecture that separates question generation from question validation.
\begin{figure*}
\centering
\includegraphics[width=0.9\linewidth]{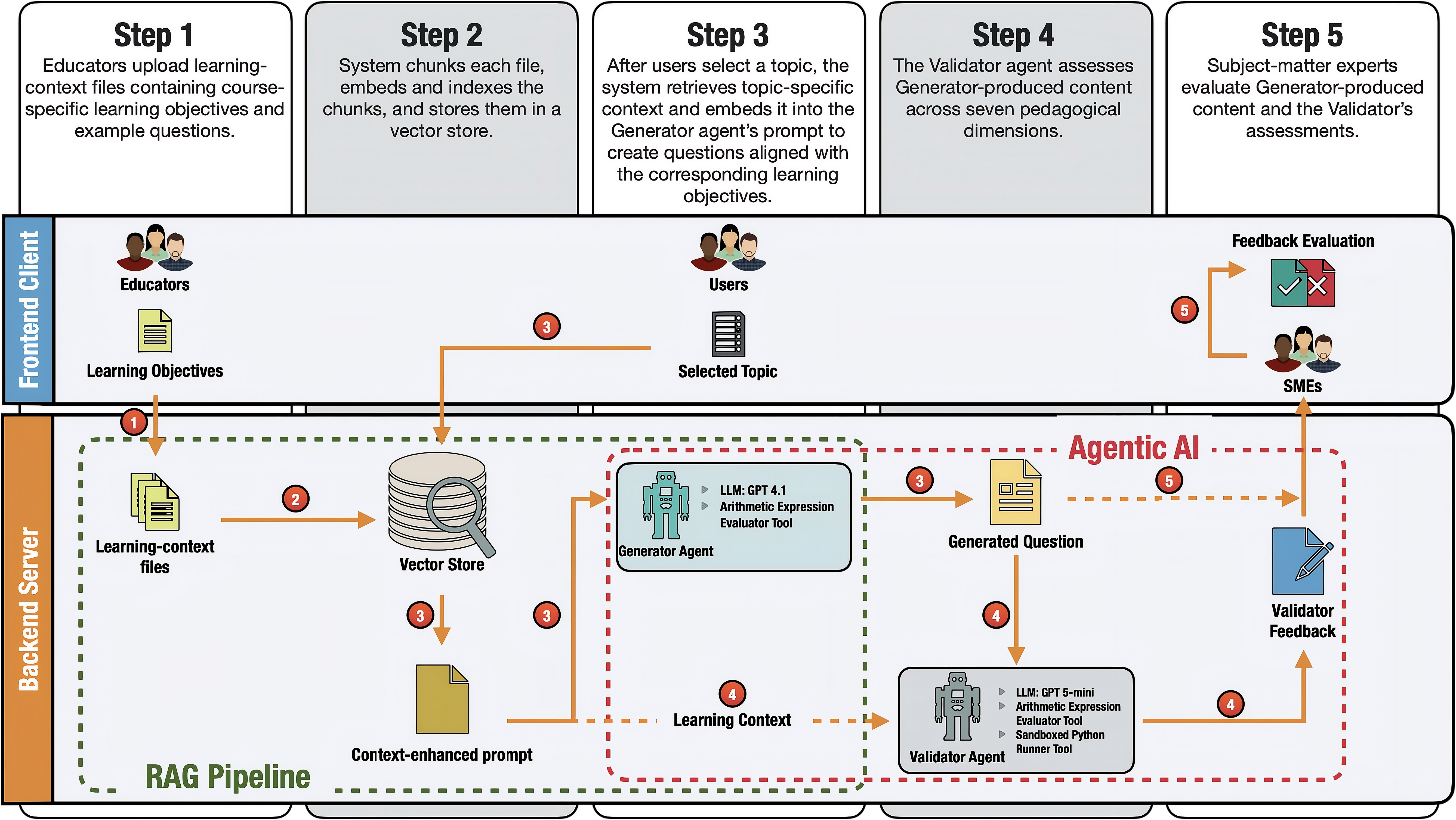}
\caption{CODE-GEN end-to-end system workflow.}
\label{system_flow}
\end{figure*}
Figure \ref{system_flow} illustrates the end-to-end workflow for CODE-GEN. The workflow begins with instructors uploading instructional materials, including learning objectives, example questions, and supporting code. These materials define the authoritative pedagogical context and are indexed through an RAG pipeline. When a user selects a topic, relevant instructional examples are retrieved and provided to a Generator agent, which produces multiple-choice questions grounded in the provided context. A Validator agent then assesses each generated question across seven pedagogical dimensions derived from established multiple-choice item-writing guidelines. Both the generated questions and the Validator’s dimension-level assessments are subsequently reviewed by SMEs, whose judgments serve as ground truth for evaluating CODE-GEN's performance.

This design explicitly treats automated validation as an empirical object of study rather than an assumed capability. By comparing Validator assessments against human expert judgments, CODE-GEN enables systematic analysis of generation patterns and targeted identification of improvement opportunities while maintaining quality assurance across large question banks.

\subsection{RAG Pipeline}
To ensure the generated questions are contextually grounded and aligned with course-specific learning objectives, CODE-GEN employs an RAG pipeline tailored to coding comprehension for multiple-choice item-writing. Uploaded instructional materials are first parsed into structured representations and segmented using a domain-specific chunking strategy that preserves semantic coherence. Unlike generic token-based splitters, this approach performs line-by-line parsing with docstring recognition, explicitly preserving learning objectives, sample questions with answer options, and associated Python code examples.

Each semantically coherent chunk is embedded using the \textit{OpenAI text embedding 3 small} model and indexed in a Facebook AI Similarity Search (FAISS) vector store configured with L2 distance metrics \cite{johnsonBillionscaleSimilaritySearch2017}. When a user initiates question generation for a given topic, the system performs nearest-neighbor retrieval using FAISS’s IndexFlatL2 algorithm to identify the most relevant instructional examples. These retrieved examples are injected directly into the Generator agent’s prompt, constraining generation to the intended instructional context and reducing the likelihood of generic or misaligned questions.

\subsection{Agentic Architecture}
At the core of CODE-GEN is an agentic AI architecture composed of two specialized agents operating sequentially: a Generator and a Validator. This separation of roles allows the system to decouple the creative task of question generation from the analytical task of quality evaluation.

The Generator agent produces multiple-choice coding comprehension questions grounded in the retrieved instructional context. Each generated question includes a stem, executable code when appropriate, four answer options, and explanatory feedback for both correct and incorrect responses. To mitigate known weaknesses of LLMs in arithmetic reasoning, the Generator is augmented with an Arithmetic Expression Evaluator tool that deterministically computes multi-step arithmetic expressions used in questions or answers.

The Validator agent independently evaluates each generated question item across seven pedagogical dimensions as described in Table \ref{validator_eval_metrix}. The dimensions are: question stem clarity \cite{haladynaReviewMultipleChoiceItemWriting2002}, code validity \cite{doughtyComparativeStudyAIGenerated2024}, concept alignment \cite{butlerMultiplechoiceTestingEducation2018}, correct answer validity \cite{townsGuideDevelopingHighQuality2014}, distractor quality \cite{gierlDevelopingAnalyzingUsing2017},  correct answer feedback quality \cite{haladynaReviewMultipleChoiceItemWriting2002}, and distractor feedback quality \cite{gierlDevelopingAnalyzingUsing2017}. To support reliable validation, the Validator employs two tools. First, it uses the Arithmetic Expression Evaluator to audit numerical claims. Second, it uses a Sandboxed Python Runner to safely execute code snippets and verify outputs or variable states in a restricted environment. This combination enables automated verification of both computational and code tracing correctness.
\begin{table}[!h]
\centering
\caption{Evaluation dimensions and Validator assessment outputs.}
{\fontsize{9pt}{9pt}\selectfont
\setlength{\tabcolsep}{3.0pt}
\renewcommand{\arraystretch}{1.2}
%{\footnotesize

\begin{tabularx}{\textwidth}{
|>{\raggedright\arraybackslash}p{2cm}
|>{\raggedright\arraybackslash}X
|>{\centering\arraybackslash}p{2.2cm}
|>{\raggedright\arraybackslash}X|}
\hline
\textbf{Evaluation Dimension} & \textbf{Description} & \textbf{Classification Output} & \textbf{Rationale Output} \\
\hline
Question Stem Clarity &
Assesses whether the question stem is clearly worded and appropriately structured (containing code or purely conceptual) &
Yes/No &
Explanation of why the question stem is or is not clear \\
\hline
Code Validity &
Verifies that the generated Python code is syntactically correct and appropriately demonstrates the target concept &
Yes/No &
Explanation of why the code is or is not valid and concept-appropriate \\
\hline
Concept Alignment &
Evaluates whether the generated question assesses the learning objective specified in the provided context &
Yes/No &
Explanation of why the question does or does not align with the specified learning objective \\
\hline
Correct Answer Validity &
Verifies whether the option marked as correct is indeed valid &
Yes/No &
Explanation of why the marked correct answer is or is not valid \\
\hline
Distractor Quality &
Assesses whether incorrect options are plausible, pedagogically meaningful, and target common student misconceptions &
Good/Poor &
Explanation of why the distractor quality is good or poor \\
\hline
Correct Answer Feedback Quality &
Evaluates whether the feedback clearly explains why the answer is correct and reinforces the underlying concept &
Good/Poor &
Explanation of why the correct answer feedback quality is good or poor \\
\hline
Distractor Feedback Quality &
Evaluates whether the feedback clearly explains why each option is incorrect and addresses the underlying misconception &
Good/Poor &
Explanation of why the distractor feedback quality is good or poor \\
\hline
\end{tabularx}
}
\label{validator_eval_metrix}
\end{table}

This agentic design provides three advantages over single-model approaches. First, it enables role-specific optimization, allowing the Generator and Validator to be tuned independently. Second, it produces transparent, dimension-level quality signals rather than opaque aggregate scores, facilitating fine-grained analysis. Third, it supports scalable quality assurance by automatically validating every generated item prior to human review.

\subsubsection{Model Selection}
To select an appropriate model for the Generator agent, we conducted comparative experiments across several commercial LLMs considered state-of-the-art at the time of the study. We focused on models accessible through APIs to ensure reproducibility and practical deployability, as open-source alternatives often require substantial computational resources and engineering effort. 

The evaluated models included Claude Sonnet 4.5, Gemini 2.5 Pro, GPT-5-mini, and GPT-4.1. Model selection criteria focused on three factors relevant to interactive educational content generation: novelty relative to retrieved context, correctness of designated answers, and response latency. Based on these comparisons, GPT-4.1 was selected for the Generator agent due to its balance of accuracy, novelty, and efficiency.

For the Validator agent, the same candidate set of models was evaluated on their ability to detect common quality issues such as incorrect answer keys and inconsistencies between answer options and feedback. GPT-5-mini demonstrated the most reliable error detection, producing concise validation analyses, and was therefore selected.

\subsubsection{Tool Augmentation}
The practice of augmenting AI agents with external functions for specialized tasks has emerged as an effective means of extending agent capabilities beyond pure language generation \cite{mastermanLandscapeEmergingAI2024}. In CODE-GEN, we adopt this approach to address well-documented limitations of LLMs in arithmetic reasoning and program execution. When performing multi-step arithmetic computations, LLMs frequently miscalculate expressions that require strict adherence to operator precedence rules, such as $2 + 3 \times (4^2) - 8 / 2$. To mitigate this limitation, we implemented an Arithmetic Expression Evaluator and integrated it into both the Generator and Validator agent workflows. When a generated question involves arithmetic reasoning, both the Generator and Validator can invoke this tool to deterministically evaluate the expression according to mathematical operator precedence rules and return an accurate result. 

We also implemented a Sandboxed Python Runner to address a core validation challenge in programming assessments: verifying the correctness of code snippets and their expected behavior without manual execution. The Validator invokes this tool to execute code within a restricted environment, capture program outputs, and inspect variable states. Together, these tools enable systematic, automated quality assurance while preserving operational security. 

\subsubsection{Prompt Engineering}
The Generator and Validator agents use structured prompts designed to ensure pedagogical alignment, reliable tool usage, and consistent output formatting. Prompts define explicit agent roles, task specifications, evaluation criteria, and tool-invocation rules. Both agents must produce outputs that conform to a predefined JSON schema, enabling consistent parsing and scalable downstream analysis. 

\subsection{Human Subject-Matter Expert Evaluation}
\begin{figure*}
\centering
\includegraphics[width=0.9\linewidth]{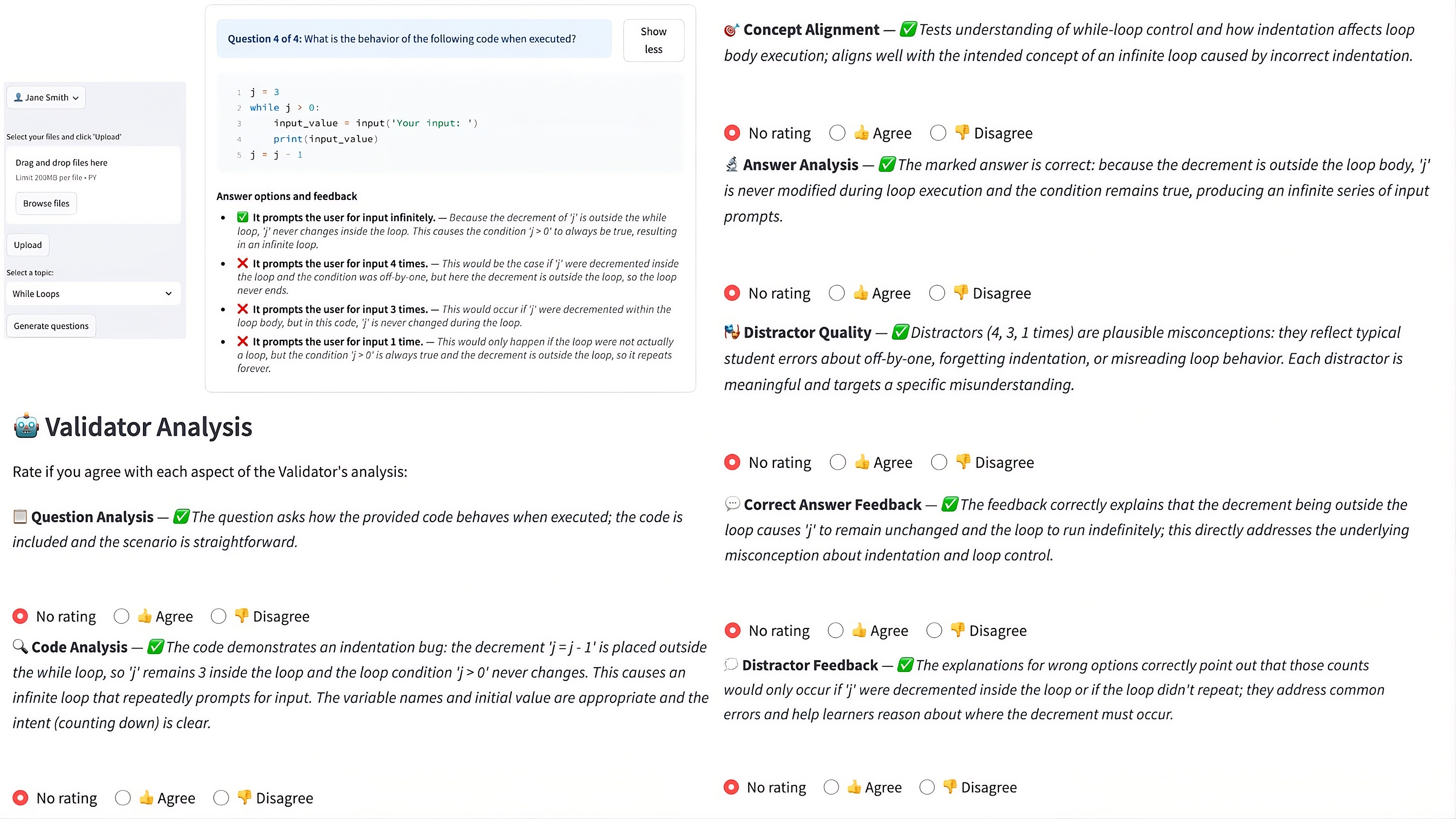}
\caption{CODE-GEN web user interface.}
\label{system_ui}
\end{figure*}

CODE-GEN leverages human expertise to verify the Validator’s assessments and mitigate the risk of propagating incorrect or misleading feedback during iterative refinement cycles. We recruited six SMEs (three men and three women), all with extensive experience teaching introductory programming. Following an orientation session to establish a shared understanding of evaluation criteria, SMEs evaluated 288 questions and the associated validations using CODE-GEN. Figure \ref{system_ui} illustrates an example of the evaluation interface using a while-loop-related question created by the Generator, including the question stem, code snippet, answer options, and explanatory feedback for each option. Adjacent to the generated content, the user interface displays the Validator’s analysis across all seven evaluation dimensions. 
For each question, SMEs judged the generated items alongside the Validator’s assessments, indicated agreement or disagreement for each dimension using the explicit Agree or Disagree button, and provided textual rationales when they disagreed. Across all experts and dimensions, this process yielded 2,016 SME-Validator rated pairs and 131 instances of qualitative feedback. 

Because CODE-GEN generates unique questions for each user, SMEs evaluated distinct question sets, which means traditional inter-rater reliability metrics (e.g., Cohen's kappa) are not applicable because they require multiple raters judging the same items. To assess evaluator consistency, we examined the SME-Validator agreement rates across all SMEs. As reported in Table \ref{tab:evaluator_consistency}, the results show consistent agreement rates across all dimensions (82.5\%--98.4\%), with five of seven exhibiting standard deviations of 3.8\% or below, indicating that SMEs applied comparable evaluation standards despite reviewing different question sets.

\begin{table}[t]
\centering
\caption{Mean and Standard Deviation (SD) of SME–Validator agreement rates.}
\label{tab:evaluator_consistency}
{\fontsize{7pt}{8pt}\selectfont
\setlength{\tabcolsep}{2.5pt}
\renewcommand{\arraystretch}{1.2}
\begin{tabularx}{\columnwidth}{|l|X|X|X|X|X|X|X|}
\hline
 & \textbf{Question Stem Clarity} & \textbf{Code Validity} & \textbf{Concept Alignment} & \textbf{Correct Answer Validity} & \textbf{Distractor Quality} & \textbf{Correct Answer Feedback Quality} & \textbf{Distractor Feedback Quality} \\
\hline
\textbf{Mean (\%)} & 98.1 & 95.3 & 98.4 & 94.7 & 82.5 & 95.3 & 89.5 \\
\hline
\textbf{SD (\%)} & 3.3 & 3.5 & 3.8 & 6.2 & 7.3 & 3.8 & 7.4 \\
\hline
\end{tabularx}
}
\end{table}

\section{Results}
\subsection{System-Level Quality Analysis}
\begin{figure*}[!h]
\centering
\includegraphics[width=0.9\linewidth]{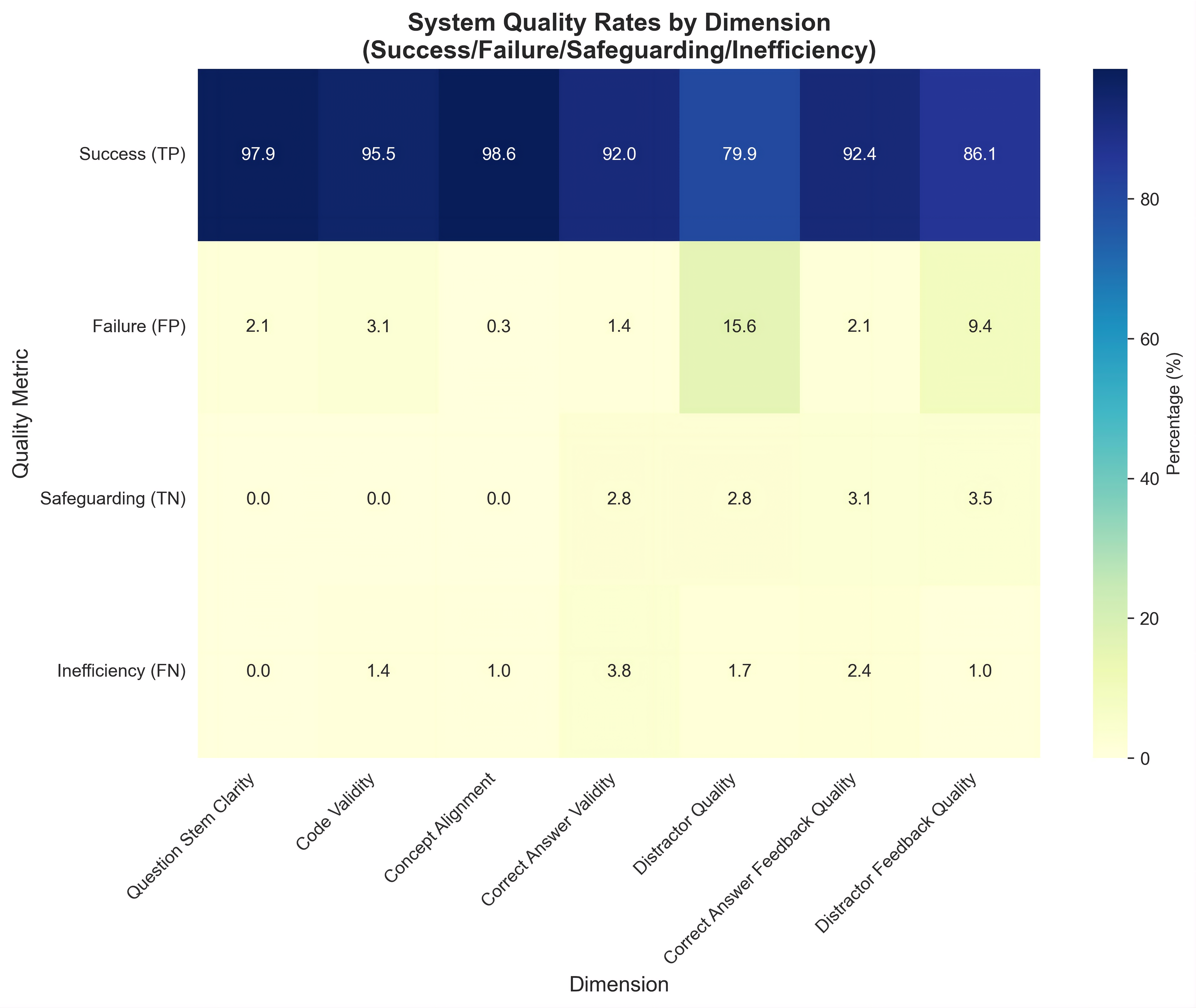}
\caption{System quality by evaluation dimension.}
\label{sys_quality_dimension}
\end{figure*}

To evaluate CODE-GEN's performance, we treated human SMEs' judgments as ground truth and the Validator's assessments as predictions in a binary classification framework. Each SME-Validator rated pair was categorized as a true positive (\textbf{TP}), false positive (\textbf{FP}), true negative (\textbf{TN}), or false negative (\textbf{FN}), as defined in Table \ref{tab:human_ai_pair_category}. We report the system performance across the seven evaluation dimensions in Figure \ref{sys_quality_dimension}.
\begin{table*}
\centering
\caption{SME-Validator rated pair category definitions.}
\label{tab:human_ai_pair_category}
{\fontsize{9pt}{9pt}\selectfont
\setlength{\tabcolsep}{3.0pt}
\renewcommand{\arraystretch}{1.2}

\begin{tabularx}{\columnwidth}{|l| X| X| c|}
\hline
\textbf{Category} & \textbf{SME vs. Validator} & \textbf{Interpretation} &  \textbf{Meaning} \\
\hline
True positive (\textbf{TP}) & SME agreed with Validator positive assessment & High-quality content correctly accepted & \textbf{Success} \\
\hline
False positive (\textbf{FP}) & SME disagreed with Validator positive assessment & Low-quality content incorrectly accepted & \textbf{Failure} \\
\hline
True negative (\textbf{TN}) & SME agreed with Validator negative assessment & Low-quality content correctly rejected & \textbf{Safeguarding} \\
\hline
False negative (\textbf{FN}) & SME disagreed with Validator negative assessment & High-quality content incorrectly rejected & \textbf{Inefficiency} \\
\hline
\end{tabularx}
}
\end{table*}

Overall, CODE-GEN demonstrates strong system performance. Five of seven dimensions achieved human-validated success rates exceeding 91\%, accompanied by minimal failure rates of 3.1\% or below. Concept alignment exhibited the highest success rate at 98.6\%, with only 0.3\%  failure and 1.0\% inefficiency. This exceptional performance validates that the RAG-based agentic design enables the Generator to produce highly targeted questions and simultaneously provides the Validator with explicit criteria for assessing alignment. 
Question stem clarity achieved 97.9\% success rate, with 2.1\% failure and no inefficiency. Code validity demonstrated 95.5\% success rate, with 3.1\% failure and 1.4\% inefficiency. These results indicate that tool augmentation substantially enhances both generation quality and validation reliability for technically grounded dimensions. Correct answer validity (92.0\% success, 1.4\% failure) and correct answer feedback quality (92.4\% success, 2.1\% failure) maintained robust success rates. They exhibited safeguarding rates of 2.8\% and 3.1\%, respectively. These results indicate that quality issues were infrequent, and when present, the Validator agent successfully identified them. 

In contrast, dimensions that require nuanced pedagogical judgment achieved comparatively lower success rates. Distractor quality yielded the lowest success rate (79.9\%) and the highest failure rate (15.6\%), while distractor feedback quality showed a success rate of 86.1\% with a comparatively elevated failure rate (9.4\%). These findings highlight a fundamental challenge in automated pedagogical evaluation: assessing whether distractors effectively target common misconceptions and whether feedback adequately elaborates on underlying concepts. This reinforces the continued need for human oversight in evaluating these dimensions.

\subsection{System-Level Limitation Analysis}
To better understand the limitations of CODE-GEN, we analyzed qualitative feedback from SMEs in FP and FN cases. This analysis reveals a systematic distinction between computational proficiency and pedagogical judgment. 

FP cases predominantly reflected limitations in pedagogical judgment rather than technical correctness. In the distractor quality dimension, the Generator often produced distractors that were syntactically valid and superficially plausible but failed to target common student misconceptions. SMEs consistently noted that higher-quality distractors would better reflect known novice confusions. For example, when the Validator approved low-quality distractors, one SME critiqued: \textit{``These distractors are obviously wrong. Better examples include pop('Leprechaun'), remove(4), and remove(3). They reflect students' common confusion with whether to use index position or values in .pop() vs. .remove() and miscount the index position by 1.''} Similarly, FP cases in feedback-related dimensions showed that the Validator frequently approved explanations that correctly described surface-level mechanics but lacked deeper pedagogical elaboration, such as explicitly connecting code behavior to underlying concepts. For example, when the Validator approved low-quality feedback, an SME noted: \textit{``The feedback didn't explain the given while loop reiterates after printing 'stop' because it reaches the end of the current iteration.''}

FN cases exposed structural weaknesses in the Validator’s reasoning process. A common pattern involved misinterpretation of answer schemas, where the Validator correctly analyzed the code output but confused the answer value with the option position. For example, when the correct answer was 2, the Validator correctly derived 2 but identified the 2nd option as marked correct when the 1st option was actually the designated correct answer. An SME noted \textit{``The marked answer is the value 2 (option 1), not option 2 (value 3).''} Another recurring issue was internal inconsistency, in which the Validator’s detailed textual analysis affirmed alignment or correctness while its final binary classification indicated the opposite. For example, the Validator's analysis explicitly affirmed concept alignment yet generated negative classifications. An SME commented, \textit{``The question aligns with the concept as stated above, but the final evaluation shows disagreement, which doesn't make any sense.''}

Taken together, these findings clarify the division of labor between AI and human expertise in educational content generation. Automated agents, when supported by retrieval grounding and computational tools, can reliably handle dimensions grounded in explicit criteria and verifiable correctness. However, dimensions requiring instructional insight, such as designing effective distractors and pedagogically rich feedback, continue to demand human judgment. These results directly inform the human-in-the-loop design of CODE-GEN and motivate the implications discussed in the following section.

\section{Discussion and Future Work}
\label{discussion}
The empirical results show that CODE-GEN achieves strong human-validated success on dimensions grounded in explicit criteria and computational verification, with success rates ranging from 79.9\% to 98.6\%. In particular, the RAG pipeline played a central role in ensuring alignment between generated questions and course-specific learning objectives, as reflected in the exceptionally high concept alignment success rate. The agentic separation between generation and validation enabled systematic, dimension-level quality control, while tool augmentation substantially improved reliability on technically grounded dimensions, such as code validity, correct answer validity, and correct answer feedback quality.

At the same time, the analysis of human-AI disagreement cases highlights important limitations of AI-assisted educational content generation. Dimensions that require deeper pedagogical judgment, including the design of effective distractors and the provision of pedagogically rich feedback, consistently benefited from human expertise. Automated validation was prone to approving assessment items that were technically correct but instructionally shallow, and it exhibited reasoning inconsistencies when interpreting structured answer representations. These findings reinforce the necessity of maintaining human oversight for aspects that depend on instructional insight and anticipation of student misconceptions.

Taken together, the results support a principled division of labor between AI and human expertise in the generation of educational content. Agentic AI systems such as CODE-GEN can provide scalable, reliable first-line quality control for dimensions involving computational verification and explicit criteria matching. However, human experts remain essential for ensuring pedagogical depth, meaningful misconception targeting, and overall instructional validity. By explicitly evaluating automated validation against human judgment, this work advances a human-centered perspective on AI-assisted educational content generation and provides design guidance for future educational AI systems.

While this study focused on introductory Python programming, the underlying architectural principles are not domain-specific. Future work should explore extending CODE-GEN to other subject areas to examine how the system performs. Additionally, the structured human expert feedback collected in this study constitutes a high-quality dataset for pedagogical alignment that can support future research on reinforcement learning from human feedback for educational LLMs. More broadly, the evaluation framework introduced in this work offers a pathway for systematically studying not only what AI systems can generate, but how reliably they can be trusted to assess quality in authentic educational settings.

\section{Conclusions}
This study presents CODE-GEN, an agentic AI system designed to 
generate contextually grounded multiple-choice coding comprehension questions. Through a systematic evaluation involving human SMEs, we demonstrated that CODE-GEN can reliably generate and automatically validate multiple-choice coding comprehension questions across a range of pedagogical dimensions, while also revealing clear boundaries of effective automation.
\begin{credits}
\subsubsection{\ackname} We gratefully acknowledge the six subject matter experts whose rigorous evaluations provided the empirical foundation for this research.

\subsubsection{\discintname}
The authors have no competing interests to declare that are
relevant to the content of this article.
\end{credits}
\bibliographystyle{splncs04}
\bibliography{refs}





% LNCS Homepage, \url{http://www.springer.com/lncs}, last accessed 2023/10/25
% \end{thebibliography}
\end{document}